\documentclass{article}
\usepackage{amsmath}
\usepackage{graphicx}
\usepackage{subcaption}
\usepackage[ruled,linesnumbered]{algorithm2e}
\usepackage{multirow}
\usepackage{comment}
\usepackage{color}
\usepackage[square,numbers]{natbib}
\usepackage{hyperref} 
\usepackage{url}
\usepackage{geometry}
\newgeometry{
textheight=9in,
textwidth=5.5in,
top=1in,
headheight=12pt,
headsep=25pt,
footskip=30pt
}
\title{Learning to Learn: Meta-Critic Networks\\ for Sample Efficient Learning}

\author{
Flood Sung\\
Independent Researcher\\
\texttt{floodsung@gmail.com}
\and
Li Zhang\\
Queen Mary University of London\\
\texttt{david.lizhang@qmul.ac.uk}
\and
Tao Xiang\\
Queen Mary University of London\\
\texttt{t.xiang@qmul.ac.uk}
\and
Timothy Hospedales\\
The University of Edinburgh\\
\texttt{t.hospedales@ed.ac.uk}
\and
Yongxin Yang\\
Yang's Accounting Consultancy Ltd\\
\texttt{yongxin@yang.ac}
}
\date{}
\begin{document}
\maketitle

\begin{abstract}
We propose a novel and flexible approach to meta-learning for learning-to-learn from only a few examples. Our framework is motivated by actor-critic reinforcement learning, but can be applied to both reinforcement and supervised learning. The key idea is to learn a meta-critic: an action-value function neural network that learns to criticise any actor trying to solve any specified task. For supervised learning, this corresponds to the novel idea of a trainable task-parametrised loss generator. This meta-critic approach provides a route to knowledge transfer that can flexibly deal with few-shot and semi-supervised conditions for both reinforcement and supervised learning. Promising results are shown on both reinforcement and supervised learning problems.
\end{abstract}


\section{Introduction}

Learning effectively from few examples is a key capability of humans that machine learning has yet to replicate, despite a long history of research \cite{Thrun1998learnToLearn,naik1992learnOptimiser,schmidhuber1997inductiveBias} and explosive recent interest \cite{Finn2017Model,Ravi2017Optimization,liMalik2017learnToOptimize,bertinetto2016feedForwardOneShot}. Like humans, we would like our AI systems to be able to learn new skills quickly: after only a few supervisions (supervised learning), or only a few interactions with the environment (reinforcement learning). To this end, the learning-to-learn vision is of  learners that address multiple learning problems, and eventually understand something about how problems relate, so that problem-independent information can be extracted and applied to help learn new tasks more quickly and accurately.

Motivated by this meta-learnning vision  -- and under various alternative names, including life-long learning, learning to learn, etc --  the community has studied various categories of approaches, including learning shared transferrable priors or weight initializations \cite{Finn2017Model,parisotto16_actormimic}, meta-models that generate the parameters of few-shot models \cite{vinyals2016matching,ha2017hypernet}, and learning effective transferrable optimizers \cite{naik1992learnOptimiser,liMalik2017learnToOptimize}.

In this paper we propose a novel approach to meta-learning that is instead based on the completely different perspective of learning a global `meta-critic' network that can be used to train multiple `actor' networks to solve specific problems. This approach is inspired by actor-critic networks in reinforcement learning (RL), where an effective strategy is to jointly train a pair of networks for any given problem such that the actor learns to solve the problem, and the critic learns how to effectively supervise the actor \cite{Barto1983Neuronlike,grondman2012actorCritic}. In RL our shared meta-critic provides the transferrable knowledge that allows actors to be trained with only a few trials on a new problem. We show that in supervised learning (SL) the same strategy corresponds to the idea of using a trainable and task-parameterised loss function, in contrast to typical fixed losses such as mean square error or cross-entropy. 

\begin{figure}	
\centering
\includegraphics[width=0.85\textwidth]{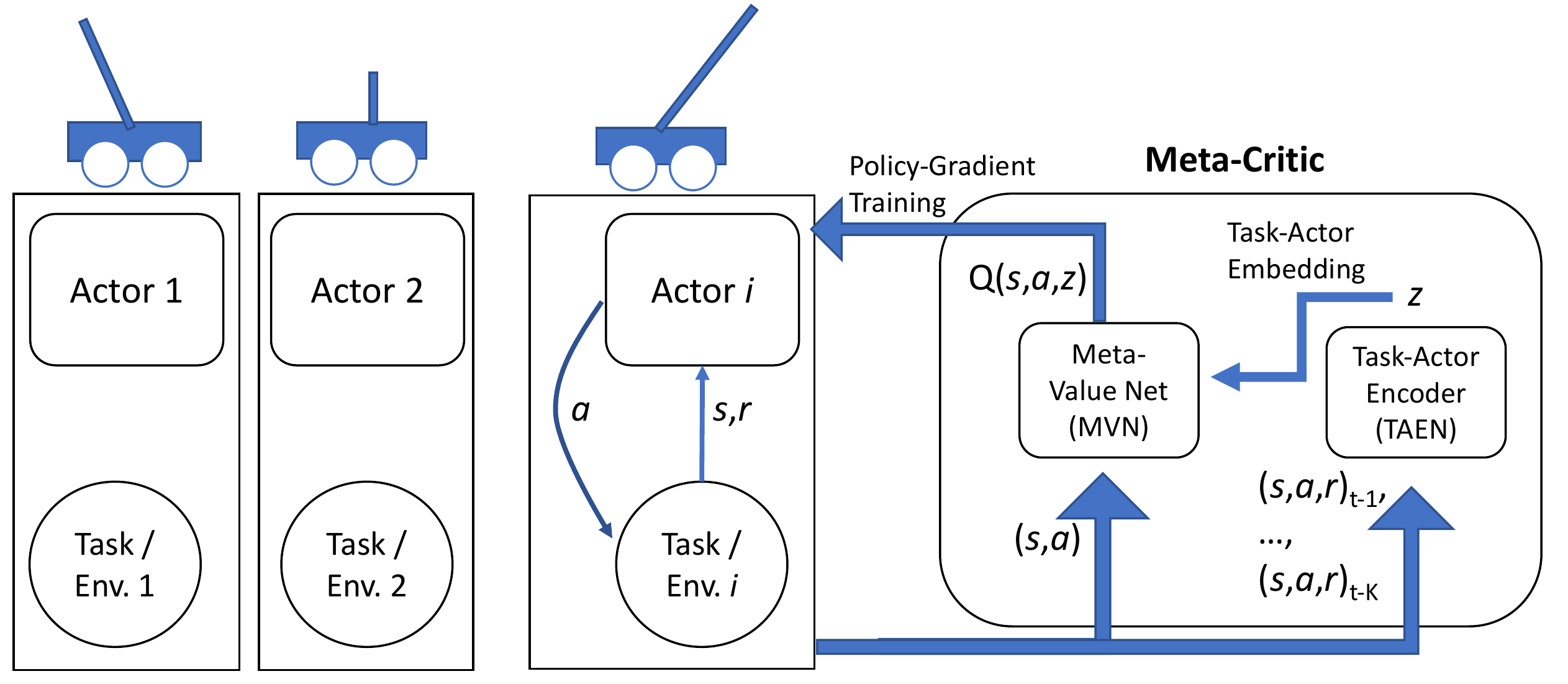}
\caption{{\small Schematic illustration of our meta-critic based approach. Meta-training: actors and meta-critic are trained. Meta-testing: Meta-critic is fixed and used to train actor for a novel task.}}
\label{fig:schematic}
\end{figure}

Conventional critic networks \cite{Barto1983Neuronlike,grondman2012actorCritic} are conditioned on a given task {and actor} that they are trained to criticise. To achieve our vision, we explicitly condition the meta-critic on an actor and a task, so that at any moment it knows what student (actor) it is training, what problem (task) the actor should be trained to solve. To achieve this explicit conditioning, we introduce the idea of a task-actor encoder network. This encoder reads a description of the current task and state of the actor learner on that task. Based on this it produces a task-actor embedding that the critic uses as input to decide how to criticise the current actor on the current task. For input to the encoder, we use an architecture agnostic featurisation of actor and task: the N-shots of inputs, labels and predictions in the case of SL; and N-trials of states, actions, and rewards in the case of RL. Fig.~\ref{fig:schematic} provides a schematic illustration of the framework. For notational simplicity we will use RL terminology and refer to the task-specific networks as actors and supervisory networks as critics for both RL and SL problems.

Our problem statement is to take a set of tasks $\mathcal{T}=\{\tau_i\}$ (each defined by feature vectors and labels in the case of SL, or POMDPs and reward functions in the case of RL) plus a new target task $\tau^*$ with few examples (SL) or allowed environmental interactions (RL) and efficiently learn a predictor/policy (SL/RL) for the target task $\tau^*$  by leveraging the background tasks $\mathcal{T}$. In contrast to other meta-learning studies  \cite{Finn2017Model,Ravi2017Optimization} addressing this problem, we explore an alternative solution of  knowledge transfer by learning a task-independent meta-critic from the source tasks that can be directly applied to training an actor for a target task. 

To understand why the meta-critic approach is effective in RL, consider that if the meta-critic can correctly criticise a new task based on the provided task-encoding, then from the perspective of the new task's actor, it benefits from a pre-trained critic which increases learning speed. Moreover, as the meta-critic is actor-conditional, it never gets `out of date' and needing to `catch-up'  with its actor, as can happen during actor critic co-evolution in conventional actor-critic architectures. The approach has a number of further benefits: It can address both SL and RL with a single framework; and both continuous and discrete outputs (classification/regression and discrete/continuous action RL). Unlike other studies, it does not make assumptions about base learner architecture like recurrent \cite{pmlr-v48-santoro16} or Siamese \cite{Koch2015SiameseNN}, or fix the number of shots used \cite{bertinetto2016feedForwardOneShot}. The ability to assume the existence of a good pre-trained critic for any target task also means that the actor can benefit from training on \emph{unlabelled} data (i.e., exploiting semi-supervised learning, or environmental interactions without access to a reward function). In each case the actor can benefit from the critic's supervision of what it should do in those unlabelled states. Finally, with a sufficiently powerful task-encoder and meta-critic, it also means that the framework is robust to diverse relatedness among the tasks -- a situation which existing approaches that search for a single shared prior \cite{Finn2017Model} are not robust to.

\section{Related Work}

\noindent\textbf{Few-shot learning}\quad There is extensive work on meta-learning to enable few-shot supervised  \cite{vinyals2016matching,Ravi2017Optimization,Koch2015SiameseNN,lake2015ppi,bertinetto2016feedForwardOneShot} and reinforcement  \cite{duan2016fastSlowRL,taylor2009TL_RL_Survey,zhao2017controlTransferTensor,schmidhuber1997inductiveBias} learning. Only a few studies provide frameworks that do not require specific model architectures and that address both settings. The most related to ours in terms of architecture agnostic few-shot learning in both SL and RL is \cite{Finn2017Model}. However, the methodologies are completely different: \cite{Finn2017Model}  aims to achieve few-shot learning by learning a shared initialisation from which any individual specific task can easily be fine-tuned. As we will show later, this approach is weak to diversity among the tasks. When learning neural networks with SGD, such a single shared initialisation roughly corresponds to a shared prior representing the `average' source task. The weights representing the average of previous tasks can be particularly bad when the distribution of tasks is multi-modal. Our meta-critic approach  is more robust to such task diversity.

\noindent\textbf{Task-parametrised Networks}\quad Our meta-critic is task-agnostic due to being parametrised by the task (and actor) to criticise. This is related to a number of topics. \textbf{Parametrised Tasks:} In learning for control, the \emph{contextual policy} or \emph{parametrised task} setting considers actors that are parametrised by some metadata describing the goal they should achieve (e.g., the goal state of a mobile agent or robot arm to reach) \cite{silva2012paramSkill,kupcsik2013optionGeneralization,silver2015uvfa}. In our case it is the critic that is task-parametrised. And rather than parameterising in terms of a high-level goal, it is parameterised by encoding the experiences of the actor. This approach simultaneously encodes both the task (by the rewards) and the actor (via the actions chosen in each state), so the critic can learn to criticise any actor for any task. \textbf{Weight Synthesis Networks:} Another group of approaches focus on designing a meta-network that generates the weights of a task-specific network based on some encoding of that task \cite{vinyals2016matching,ha2017hypernet,larochelle2008zeroData,bertinetto2016feedForwardOneShot}, so that the task-agnostic knowledge is encoded in the weights of the meta-network. In our case the task-agnostic knowledge is stored in the task-encoder and critic networks, and we use the task-encoding to drive the supervision of the target network via the critic, rather than synthesise its weights. Our `learning to supervise' rather than this `learning to synthesise' approach has several advantages: it allows application to RL, which is not non-trivial for existing weight-synthesis methods oriented at SL; it allows learning from unlabelled data / unrewarded interactions with the environment \cite{finn2017sslRL}; and it does not constrain the meta and target networks to have matching architectures.

\textbf{Distillation}\quad Our approach is somewhat related to knowledge distillation  \cite{bucila2006modelCompression,hinton2014distilling} -- the supervision of a student network by a teacher network. The meta-critic shares distillation's favourable ability to use unlabelled data to train the student actor. In contrast to regular distillation where teacher and student are both single-task actors, our teacher is a meta-critic shared across all tasks. Related applications of distillation include actor-mimic \cite{parisotto16_actormimic} where multiple expert actor networks teach a single student actor how to solve multiple tasks. This is the opposite of our single teacher critic teaching multiple actor networks to solve different tasks (and thus providing knowledge transfer).

\textbf{Multi-Actor Methods}\quad Multiple actors with a centralised critic has been used in multi-agent learning context for decentralised training of centralised policies \cite{Foerster2017Counterfactual}. While A3C \cite{Mnih2016Asynchronous} uses multiple actors to generate diverse uncorrelated samples of the environment to improve learning speed. These studies address actors learning to solve the \emph{same} task independently or cooperatively, rather than our actors addressing distinct tasks, and meta-critic that generalises across multiple tasks.

\section{Methodology}

We first introduce our methodology from a continuous reinforcement learning perspective, and later explain the discrete action (Section~\ref{sec:mvn_da_rl}) and supervised learning (Section~\ref{sec:mvn_sl}) variants. 

\noindent\textbf{Background}\quad In the RL problem, at each  time $t$, an agent receives a state $s_t$, based on which it takes an action $a_t$, after that the agent will observe a new state $s_{t+1}$ and reward $r_t$, according to the dynamics of the environment. We focus on model-free reinforcement learning, so the agent does assume a model of the transition: $(s_t, a_t) \rightarrow s_{t+1}$, nor reward: $(s_t, a_t, s_{t+1}) \rightarrow r_t$.

Our proposed method is based on the actor-critic architecture \cite{Barto1983Neuronlike,grondman2012actorCritic}, and we use neural networks as function approximators for both actor (policy network) and critic (value network).

In continuous-action RL, the actor produces the \emph{actual} action and is trained by maximising the output of critic (i.e., deterministic policy \cite{Silver2014Deterministic, Lillicrap2016Continuous}). The critic is the state-action value function (Q-function \cite{Watkins1992Q}) and is trained \textcolor{black}{to estimate the future return} by minimising the temporal difference error\footnote{Note that we choose deterministic policy and Q-function here for illustration purpose only, and we do \emph{not} make assumptions on how actor and critic are designed except they are both realised by neural networks. E.g., some alternative choices: actor can produce a Gaussian distribution over actions (i.e., Gaussian policy) and be trained by actor-critic policy gradient \cite{Sutton1999Policy} with either Q-function or advantage function \cite{Baird1993Advantage, Mnih2016Asynchronous, Schulman2016High}.}.

The policy network (actor) is parametrised by $\theta$ and it takes as input the state $s_t$, then outputs an action $a_t$, i.e., $a_t = P_\theta (s_t)$. A value network is parametrised by $\phi$ and it takes as input the state $s_t$ and action $a_t$, then outputs an expected discounted reward, i.e., $Q^{P_\theta}_\phi (s_t, a_t) = r_t + \gamma Q^{P_\theta}_\phi (s_{t+1}, a_{t+1})$. Where the notation $Q^{P_\theta}$ indicates the critic is trained to model the return of policy $P_\theta$. 

The objective of the value network is to predict expected discounted reward as accurately as possible, and the objective of the policy network is to maximise the discounted future reward, assumed to be the return estimated by the value network. Therefore the optimisation procedure is to alternatively update policy network and value network:
\begin{eqnarray}
&\theta \leftarrow \underset{\theta}{\operatorname{argmax}} ~~ Q^{P_\theta}_\phi (s_t, a_t)\\
&\phi \leftarrow \underset{\phi}{\operatorname{argmin}} ~~ \big(Q^{P_\theta}_\phi (s_t, a_t) - r_t - \gamma Q^{P_\theta}_\phi (s_{t+1}, a_{t+1})\big)^2 \label{eq:critic_q_func}
\end{eqnarray}
\noindent where $a_t = P_\theta (s_t)$ and $a_{t+1} = P_\theta (s_{t+1})$.

\noindent\textbf{A Meta-Critic and Task-Actor Encoder for Multiple Tasks}\quad Our vision is of a single meta-critic that can criticise \emph{any actor}, performing \emph{any task}. This requires two generalisations (task and actor conditioning) compared to conventional critic networks that criticise a specific actor for a specific task. In order for the value network to issue correct task- and actor-specific supervision, we need an additional input beyond the conventional state-action pair. We encode the task and the actor via a task-actor embedding  $z_t$. To generate $z_t$, we propose to use a task-actor encoder network (TAEN) $C_\omega$ parametrised by $\omega$. The task network inputs a featurisation of the actor and the task that it is supposed to be solving and produces the task embedding $z_t$. There are various possible featurisations (e.g., the actor's weights), but we take a simple featurisation that makes no assumption on the form of the actor. Specifically, we define the task-actor encoder as a recurrent neural network (RNN)  that inputs a \emph{learning trace} as a sequence of $k$ triplets $L^t_{t-k} = [(s_{t-k}, a_{t-k}, r_{t-k}), (s_{t-k+1}, a_{t-k+1}, r_{t-k+1}), \dots, (s_{t-1}, a_{t-1}, r_{t-1})]$ and outputs the task-actor encoding $z_t = C_\omega(L^t_{t-k})$. To understand this encoding, observe that by including  \emph{state-action} pairs the encoder sees a featurisation of the actor it is to criticise (choice of action depends on actor parameters). Meanwhile by observing the \emph{rewards} obtained, it sees a featurisation of the task the actor is solving. The meta-critic is thus composed of the meta-value network (MVN) and task-action encoder network (TAEN), and shared across all tasks and actors. Assuming $M$ training tasks, the update rules for every individual actor $i$ and meta-critic are:
\begin{eqnarray}
&\theta^{(i)} \leftarrow \underset{\theta^{(i)}}{\operatorname{argmax}} ~~ Q_\phi (s_t^{(i)}, a_t^{(i)}, z_t^{(i)}) \quad \forall i \in [1,2,\dots,M]\label{eq:mvn_a}\\
&\phi, \omega \leftarrow \underset{\phi, \omega}{\operatorname{argmin}} ~~ \sum_{i=1}^{M}\big(Q_\phi (s_t^{(i)}, a_t^{(i)}, z_t^{(i)}) - r_t^{(i)} - \gamma Q_\phi (s_{t+1}^{(i)}, a_{t+1}^{(i)}, z_{t+1}^{(i)})\big)^2\label{eq:mvn_c}
\end{eqnarray}
\noindent where $z_t^{(i)} = C_\omega({L^{(i)}}_{t-k}^{t})$ and $a^{(i)}_t=P_{\theta^i}(s^{i}_t)$, and we  drop the superscript $^{P_{\theta}}$ of $Q$ to indicate that $Q$ is now parametrised by $P_\theta$ rather than trained to fit a specific $P_\theta$.  When optimising $\theta$, \textcolor{black}{the gradient due to $z_t$ is ignored, i.e., we see $z_t$ as a constant rather than a function involving $\theta$ \cite{Andrychowicz2016}}.

\subsection{Discrete-action RL}
\label{sec:mvn_da_rl}
For discrete-action RL, we choose to use actor-critic policy gradient. The policy network $P_\theta$ outputs a categorical distribution $o_t$ over available actions. Actions are then sampled from the produced distribution $a_t\sim P_\theta(s_t)$ , where $a_t$ is the one-hot encoding vector indicating the chosen action. The value network is optimised as in  Eq.~\ref{eq:critic_q_func}, and the policy network is optimised by
\begin{equation}
\theta \leftarrow \underset{\theta}{\operatorname{argmin}} ~~ \ell(o_t, a_t) Q^{P_\theta}_\phi (s_t, a_t), \label{eq:acpg_a}
\end{equation}
\noindent where $o_t = P_\theta (s_t)$ and $\ell(\cdot, \cdot)$ is the cross entropy loss function. 
The extension from a single RL task to multiple ones is similar to the case of continuous-action RL (Eq.~\ref{eq:mvn_a} and Eq.~\ref{eq:mvn_c}) except that the policy network update (Eq.~\ref{eq:mvn_a}) is modified according to Eq.~\ref{eq:acpg_a}, leading to
\begin{equation}
\theta^{(i)} \leftarrow  \underset{\theta}{\operatorname{argmin}} ~~ \ell(o_t^{(i)}, a_t^{(i)}) Q_\phi (s_t^{(i)}, a_t^{(i)}, z_t^{(i)})  \quad \forall i \in [1,2,\dots,M]\label{eq:mvn_acpg_a}
\end{equation}

\subsection{Supervised Learning}
\label{sec:mvn_sl}

\textbf{An Actor-Critic Reframing of Supervised Learning}\quad To apply our framework to supervised learning, we imagine a  one-step only game, where the actor (function approximator to learn) takes as input a feature $x$, and outputs the estimation of target variable $\hat{y}$. Then the environment returns the reward calculated from the negative loss value, i.e., $r = -\ell(\hat{y},y)$ and the game terminates. Differently to conventional supervised learning, in this case we assume the  ground truth $y$ is  \emph{not}  exposed to the agent directly, and the agent has no access to the form of loss function. This makes it more like a zero-order (black-box) optimisation problem rather than conventional supervised learning problem. 

The objective of critic is now to predict the \emph{current} reward since the game has one step only.
\begin{eqnarray}
&\theta \leftarrow \underset{\theta}{\operatorname{argmax}} ~~ Q_\phi (x, \hat{y})\label{eq:mvn_sl_a}\\
&\phi \leftarrow \underset{\phi}{\operatorname{argmin}} ~~ \big(Q_\phi (x, \hat{y}) - r)^2\label{eq:mvn_sl_c}
\end{eqnarray}
\noindent where $\hat{y} = P_\theta (x)$ and $r = -\ell(\hat{y},y)$. We emphasise that though the reward is calculated by $-\ell(\hat{y},y)$, this calculation is hidden in the environment. Thus we can see the correspondence to the previous RL problem: (i) state $s\rightarrow x$, (ii) action $a\rightarrow \hat{y}$, and (iii) reward $r\rightarrow -\ell(\hat{y},y)$.

\textbf{Meta-Critic-based Supervised Learning}\quad 
With the above actor-critic interpretation, extending  single-task SL problem to multiple tasks is similar to the case of continuous-action RL (Eq.~\ref{eq:mvn_a} and Eq.~\ref{eq:mvn_c}) except the actor-critic updates are changed according to Eq.~\ref{eq:mvn_sl_a} and Eq.~\ref{eq:mvn_sl_c}: 
\begin{eqnarray}
&\theta^{(i)} \leftarrow \underset{\theta^{(i)}}{\operatorname{argmax}} ~~ Q_\phi (x^{(i)}, \hat{y}^{(i)}, z^{(i)}) \quad \forall i \in [1,2,\dots,M]\label{eq:mvn_a_m}\\
&\phi, \omega \leftarrow \underset{\phi, \omega}{\operatorname{argmin}} ~~ \sum_{i=1}^{M}\big(Q_\phi (x^{(i)}, \hat{y}^{(i)}, z^{(i)}) - r_t^{(i)}\big)^2\label{eq:mvn_c_m}
\end{eqnarray}
Here we can see that the function approximator (actor) is learning to maximize the negative supervised learning loss, as estimated by the meta-critic, rather than minimise a fixed loss function as in regular SL. Meanwhile the meta-critic learns to \emph{simulate the actual supervised learning loss of each problem} $i=1\dots M$. For a single-task, this indirection just makes learning more inefficient. But crucially for multiple tasks, it means the loss generator (meta-critic) benefits from cross-task knowledge sharing. And because the tasks are parametrised, it means the loss generator can immediately supervise a new task `out of the box' based on its parametrisation, without needing much data.

\textbf{Task Parametrisation}\quad As we are focused on rapid few-shot learning, we describe tasks by way of few-shot examples similarly to \cite{bertinetto2016feedForwardOneShot,vinyals2016matching}. Other parameterisations such as word vectors used in zero-shot learning are also possible \cite{larochelle2008zeroData,frome2013devise}. Specifically, we stick with an RNN-based task-encoder $C$,  that encodes the few-shot examples into a fixed length embedding  describing the task. Analogous to our  $[s,a,r]$ learning trace-based task description in RL, we would use the mirror $[x,\hat{y},-\ell(y,\hat{y})]$ for SL. However, we find that $[x,y,0]$ works  slightly better in practice. I.e., we assume, for the purpose of constructing the input for TAEN, that the actor is able to produce the perfect estimation.

\subsubsection{Seamless Integration with Semi-Supervised Learning}

In addition to  few-shot training data of $[x,y,l(y,\hat{y})]$ or $[s,a,r]$ tuples, it is often the case that we have access to a larger set of unlabelled data: $[x]$ in in the case of SL, or $[s,a]$ tuples in the case of RL. This arises when examples are plentiful but costly to annotate in SL, or when environmental interactions are cheap but the reward function is costly to provide in RL \cite{finn2017sslRL}. Crucially, a unique advantage of our approach is that the supervisory signal generated by the meta-critic does not need the ground truth $y$ after training. Similar to other meta-learning experimental designs, we assume a meta-training stage on a set of background tasks, and then a meta-testing stage where we should learn a novel task with few trials. In meta-learning stage both meta-critic and actors are trained. In meta-testing our meta-critic is fixed and only the new task's actor is trained. The few examples of the meta-testing task are used to tell the meta-critic how to supervise via the TAEN. But the actor's training can include unlabelled data from the testing task and receive supervision about those from the meta-critic instead of actual (unavailable) environmental rewards (RL) or instance labels (SL).

\section{Experiments}

\subsection{Implementation and Details}

Applying meta-learning with a focus on few-shot learning proceeds in three stages: (i) \textbf{Meta-training} on multiple source tasks; (ii) \textbf{Meta-testing} to learn a new task with few training samples (or environmental interactions); (iii) \textbf{Final-Testing} the out-of-sample performance on the new task.

In  meta-training (i) we train multiple actors for each task along with the meta-critic to supervise the actors. This process is summarised in Algorithm~\ref{algo:mvn_training}. Here the scheduling of alternating actor-critic  updates is determined by the task mini-batch size. Two extremes are: (i) when the mini-batch size is $1$, the critic (both MVN and TAEN) is updated after any actor is updated; (ii) when the mini-batch size is $M$, the critic is updated after all actors are updated. In  meta-testing  (ii) we train an actor for a new held out task. The parameters of the meta-critic and task-encoder are fixed. The few-shot data for the new-task provides the task-actor description via the TAEN, and (possibly along with unlabelled data) provides the input to the MVN, from which the meta-critic provides a supervisory signal for training the new actor on the new task. The process is summarised in Algorithm~\ref{algo:mvn_adapting}.

For all experiments, we use the following neural networks architecture: (i) actor is an MLP with $2$ hidden layers (ReLU), and the number of neurons for both hidden layers is $40$. (ii) Task-action encoder network (TAEN) is a one-layer LSTM \cite{Hochreiter1997LSTM} with $30$ cell units, and the last hidden state is connected with a dense layer to $3$ output neurons. (iii) Meta-value network (MVN) is an MLP with $2$ hidden layers (ReLU), and the number of neurons for both hidden layers is $80$. The input for the MVN is of the length $=\text{state size} + \text{action size} + \text{TAEN output size} (3)$.

\IncMargin{1.5em}
\begin{algorithm}[H]
\KwIn{Task generator $\mathcal{T}$}
\KwOut{Trained task and value net}
Init: task and value net\;
\For{episode = 1 to max episode}{
Generate $M$ tasks from $\mathcal{T}$\;
Init $M$ policy nets (actors)\;
\For{step = 1 to max steps}{
Sample mini-batch of tasks\;
\ForEach{task in mini-batch}{
Sample training data from task\;
Train task-specific actor\;
}
Train value network\;
Train task network\;
}
}
\caption{Meta-Learning Stage}
\label{algo:mvn_training}
\end{algorithm}
\DecMargin{1.5em}

\IncMargin{1.5em}
\begin{algorithm}[H]
\KwIn{An unseen task}
\KwIn{Trained task and value nets}
\KwOut{Trained policy network}
Init: one policy network (actor)\;
\For{step = 1 to max step}{
Sample train data from task\;
Train actor\;
}
\caption{Meta-Testing Stage}
\label{algo:mvn_adapting}
\end{algorithm}
\DecMargin{1.5em}

\textbf{Baselines}\quad In most experiments, we compare four methods: \textbf{Standard:} For a new task train an actor directly from scratch, as per standard SL/RL (no stage (i)). \textcolor{black}{\textbf{All+FT:} Train a single actor  on all the source tasks together (stage (i)) before fine-tuning it (with a new output layer) on the target}\footnote{The efficacy of this baseline will depend on whether the loosely termed `tasks' are really differing tasks (with different reward functions), or contexts/domains (different word models) \cite{taylor2009TL_RL_Survey}.}.  \textbf{Model-Agnostic Meta Learning (MAML):} State of the art meta-learner for learning a transferable source model \cite{Finn2017Model}. \textbf{Meta-Critic:} Our Meta-critic network as described above.

\subsection{Supervised Learning: Regression}

Inspired by \cite{Finn2017Model}, we synthesise regression problems based on sinusoidal $a\sin(x+b)$ and linear  $cx+d$ functions. Within these function classes, different tasks can be synthesised by choosing different $a$, $b$, $c$, and $d$ values. The ranges of these variables are: $a\in [1,5]$, $b\in [0,\pi]$, $c\in [-3,3]$, and $d\in [-3,3]$. The range of $x$ is in $[-5,5]$. We experiment with two conditions: the first involving sine functions only (as in \cite{Finn2017Model}), and the second one has a mixture of sine and linear functions (half-half) \textcolor{black}{in both meta-training and testing stages}.
\textcolor{black}{For meta-training (i), we synthesise $10,000$ tasks with $30,000$ samples per task.} Meta-learning thus means learning  about the function-class of sinusoids and lines. If successful, then for a meta-testing task, a meta-learner should be able to estimate the line/sinusoid parameters from a few examples, and immediately be able to make a good prediction.

For meta-testing each task has only a few-shots $K\in\{4,6,8\}$ pairs of $(x,y)$ for training. We generate $100$ new testing tasks in total, and for each generate $100$ testing samples on which to evaluate the final-testing  MSE. We repeat every adapting round (corresponding to a task) $10$ times with different $(x,y)$ pairs, so one testing task has $10$ MSE values on the \emph{same} out-of-sample set. The mean and standard deviation of all tasks' MSEs are summarised in Table~\ref{tab:mvn_sl_all}. For our method we also explore semi-supervised learning using unlabelled $x$. In real-world problems these would correspond to unlabelled instances, but for this synthetic problem, we  simply uniformly sample $x\sim\left[-5,5\right]$. We can see that our meta-critic performs comparably or better than alternatives.  Here we also evaluate Meta-Critic-SL: our method where the meta-testing actor is pre-trained with standard supervised learning before training by the meta-critic. The similar performance of these two variants shows that via the shared meta-critic, a good performing actor can be obtained by learning from scratch, without requiring any pre-training. 

\begin{table}
\centering
\resizebox{.99\textwidth}{!}{%
\begin{tabular}{c c c c c c c}
\hline
&
\multicolumn{3}{c}{Sinusoid only} &
\multicolumn{3}{c}{Sinusoid and Linear Mixture} \\\hline
Num. of Samples & 4 & 6 & 8 & 4 & 6 & 8 \\
\hline
Standard     &  13.42 (20.21) &  6.17 (11.26) &  3.55 (7.17) &          8.18 (16.83) &    3.82 (9.93) &    1.90 (6.91) \\
All+FT &    6.41 (5.02) &   6.10 (4.73) &  5.82 (4.42) &         20.61 (26.23) &  19.90 (25.46) &  19.80 (25.59) \\
MAML \cite{Finn2017Model} & 0.87 (0.98) & 0.55 (0.75) & 0.53 (0.60) & 8.99 (11.56) & 3.68 (4.69) & 2.79 (3.13) \\
Meta-Critic    &    \textbf{0.42 (0.61)} &   0.15 (0.37) &  \textbf{0.09 (0.11)} &          7.91 (18.08) &   \textbf{2.65 (11.18)} &    \textbf{0.86 (3.00)} \\
Meta-Critic-SL       &    0.45 (0.81) &   \textbf{0.14 (0.36)} &  0.11 (1.06) &          \textbf{7.69 (13.93)} &   3.06 (10.22) &    1.26 (5.98) \\
\hline
\end{tabular}%
}
\caption{{\small Final-testing performance for different regression meta learning strategies (MSE).}}
\label{tab:mvn_sl_all}
\end{table}

\begin{figure}
\centering
\includegraphics[width=0.9\textwidth]{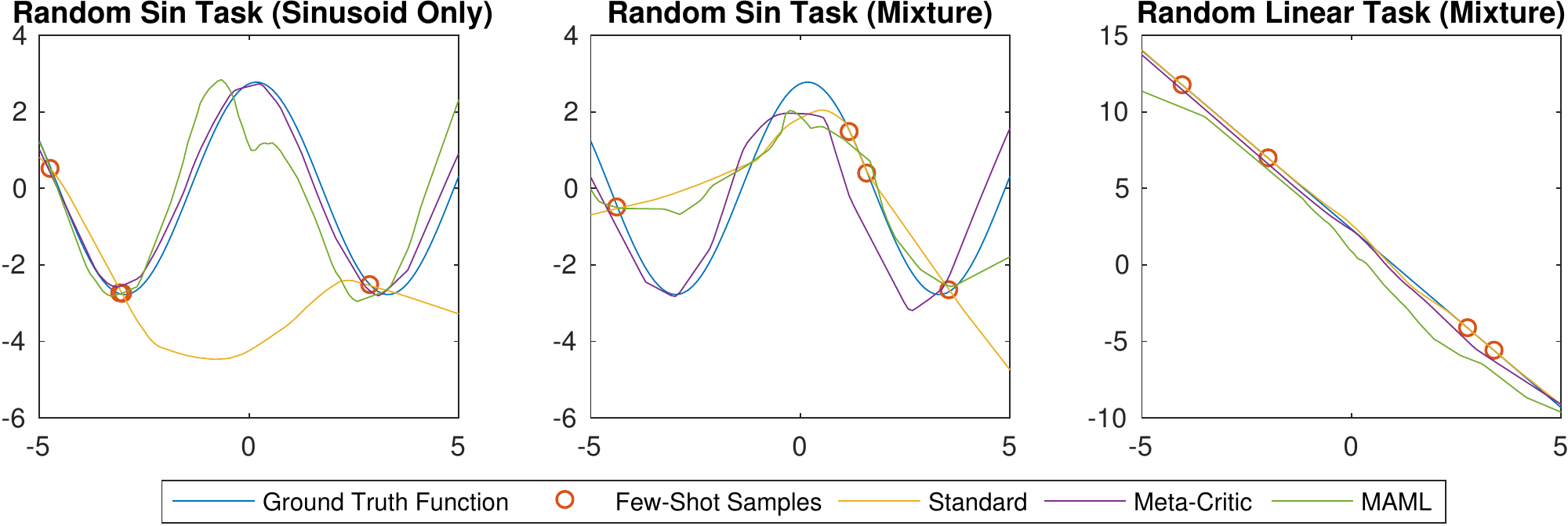}
\caption{{\small Qualitative comparison of  meta-learning strategies on few-shot learning regression tasks drawn from a function pool of sin only (Sinusoid), versus a mixture of sin and linear (Mixture).}}
\label{fig:mvn_regression}
\end{figure}

For qualitative illustration, we randomly pick tasks for \textcolor{black}{$K=4$ shot learning} from sinusoid only condition and line+sinusoid mixture condition, as shown in Fig.~\ref{fig:mvn_regression}. We see that all models fit the $K=4$ shot meta-testing data. In the sin-only condition (Fig.~\ref{fig:mvn_regression}(a)) MAML is not much worse than meta-critic in the out-of-sample areas. However, in the mixture condition (Fig.~\ref{fig:mvn_regression}(b)), MAML is much worse than meta-critic. The reason is that a single globally shared prior/initialisation parameter implicitly assumes that  tasks are similar, and their distribution is uni-modal. In the mixture case where tasks are diversely distributed and with varying relatedness, then this assumption is too strong and performance is poor. We increased the number of parameters for the actor network in MAML so that there was ample capacity for learning a complex multi-modal model, but it didn't help. In contrast, our meta-critic approach is flexible enough to model this multi-modal task-distribution and learns to supervise the actor appropriately. These qualitative results are reflected quantitatively in Table~\ref{tab:mvn_sl_all}, where we see that \textcolor{black}{MAML and Meta-Critic perform comparably -- and better than Standard/All-Fine-Tune -- in the Sin only condition, but Meta-Critic is clearly better in the mixture condition.} This is because the TAEN successfully learns to embed task category information in its task description $z$. Taking the $z$s for all the tasks in the mixture condition, we find that a simple classifier (SVM with RBF kernel) can obtain $85\%$ accuracy in predicting the task category (linear vs sinusoid). 

\subsection{Reinforcement Learning: Dependant Multi-arm Bandit}

For a first RL experiment, we work with multi-arm bandits (MAB). In this experiment, the arm rewards are dependent: the probability of getting a reward is drawn a Dirichlet distribution parametrised by $\alpha_1=\alpha_2=\dots=1$, so the sum of reward probabilities over all arms is one. Dependent MAB means that a trial on one arm also tells us about the other arms. However, it is not encoded in the model, i.e., the agent is unaware of the distribution, or that the arms are dependent. It is expected to find this out for more sample-efficient learning. Each task is a different arm reward configuration:  multiple samples drawn from the same Dirichlet distribution. We experiment with $2-$, $4-$, and $6-$arm bandits. For meta-training (i), we generated $1000$ tasks (bandits) with ample ($30,000$) samples, and for meta-testing (ii) we generate $100$ new bandits (corresponding to different arm-reward distributions) and only a few trials are allowed per task. We repeat  experiments $10$ times due to the randomness of getting a reward by pulling an arm.

The  results are shown in Table~\ref{tab:mvn_bandit_all}, quantified by average reward which is calculated by the dot product of its softmax output and the bandit's configuration (probability of getting a reward by pulling each arm). The Upper bound is calculated by always pulling the arm with largest probability of getting reward, which is $0.75$ for 2-arm, $0.52$ for 4-arm, and $0.41$ for 6-arm. The random choice lower bound is always equal to $\frac{1}{\text{Num. of Arms}}$. Our meta-critic strategy achieves higher reward than competitors for any given number of trials.

\begin{table}
\centering
\resizebox{.99\textwidth}{!}{%
\begin{tabular}{c c c c c c c c c c}
\hline
&
\multicolumn{3}{c}{$2-$arm} &
\multicolumn{3}{c}{$4-$arm} &
\multicolumn{3}{c}{$6-$arm}\\
Num. of Pulls & 5 & 10 & 15 & 10 & 15 & 20 & 15 & 20 & 25  \\
\hline
Random & 	\multicolumn{3}{c}{$0.5$} & 	\multicolumn{3}{c}{$0.25$}& 	\multicolumn{3}{c}{$0.17$} \\
Standard          &  0.69 (0.17) &  0.69 (0.17) &  0.69 (0.17) &  0.41 (0.11) &  0.43 (0.12) &  0.43 (0.12) &  0.29 (0.07) &  0.29 (0.08) &  0.29 (0.07) \\
All+FT &  0.43 (0.30) &  0.43 (0.30) &  0.43 (0.30) &  0.31 (0.22) &  0.31 (0.22) &  0.31 (0.22) &  0.13 (0.13) &  0.13 (0.13) &  0.13 (0.13) \\
MAML \cite{Finn2017Model} & 0.61 (0.25) & 0.63 (0.24) & 0.65 (0.25) & 0.35 (0.18) & 0.38 (0.22) & 0.38 (0.19) & 0.26 (0.14) & 0.27 (0.15) & 0.27 (0.15)\\
Meta-Critic & \textbf{0.70 (0.18)} &  \textbf{0.73 (0.17)} &  \textbf{0.74 (0.17)} &  \textbf{0.44 (0.14)} &  \textbf{0.47 (0.16)} &  \textbf{0.48 (0.16)} &  \textbf{0.30 (0.10)} &  \textbf{0.31 (0.11)} &  \textbf{0.32 (0.11)} \\
Best &  \multicolumn{3}{c}{$0.75$}& \multicolumn{3}{c}{$0.52$} & \multicolumn{3}{c}{$0.41$} \\
\hline
\end{tabular}%
}
\caption{{\small Different meta-learning strategies for dependent multi-arm bandit. Reward in final-testing.}}
\label{tab:mvn_bandit_all}
\end{table}

\subsection{Reinforcement Learning: Cartpole Control}

Finally we try a classic RL control problem -- cartpole. The agent should learn to balance a pole by moving a cart left or right to keep it stable \cite{gym}. \textcolor{black}{The state $s_t$ is $4$-dimensional $(\theta, x, \dot{\theta}, \dot{x})$ where $\theta$ is the pole angle, $x$ the cart position, and $(\dot{\theta}, \dot{x})$ are their respective velocities.} For meta-training, we generate $1000$ tasks by sampling pole lengths in the range $[0.5,5]$. For each task, we train for $150,000$ interactions (actor moves left or right given the state). For meta-testing, the agent is allowed to play $100$ full games (episodes), while the game terminates when the pole falls or $200$ interactions have been reached. After each game terminates, we update the model parameter using the experience collected, and start an offline testing: let the agent play $10$ games (episodes), and record the average reward (this experience will not be used for any training purpose). We carry out the experiment on $100$ new tasks, and repeat it $10$ times for each task. In this experiment, we exclude All+FT as it performed much worse than the others and include the PG-ELLA \cite{ammar2014pgella} method (lifelong  policy gradient-based  reinforcement learning via a low-rank policy factorisation), which we extended to work with a neural network policy as per \cite{zhao2017controlTransferTensor}. 

\begin{figure}
\centering
\includegraphics[width=0.70\linewidth]{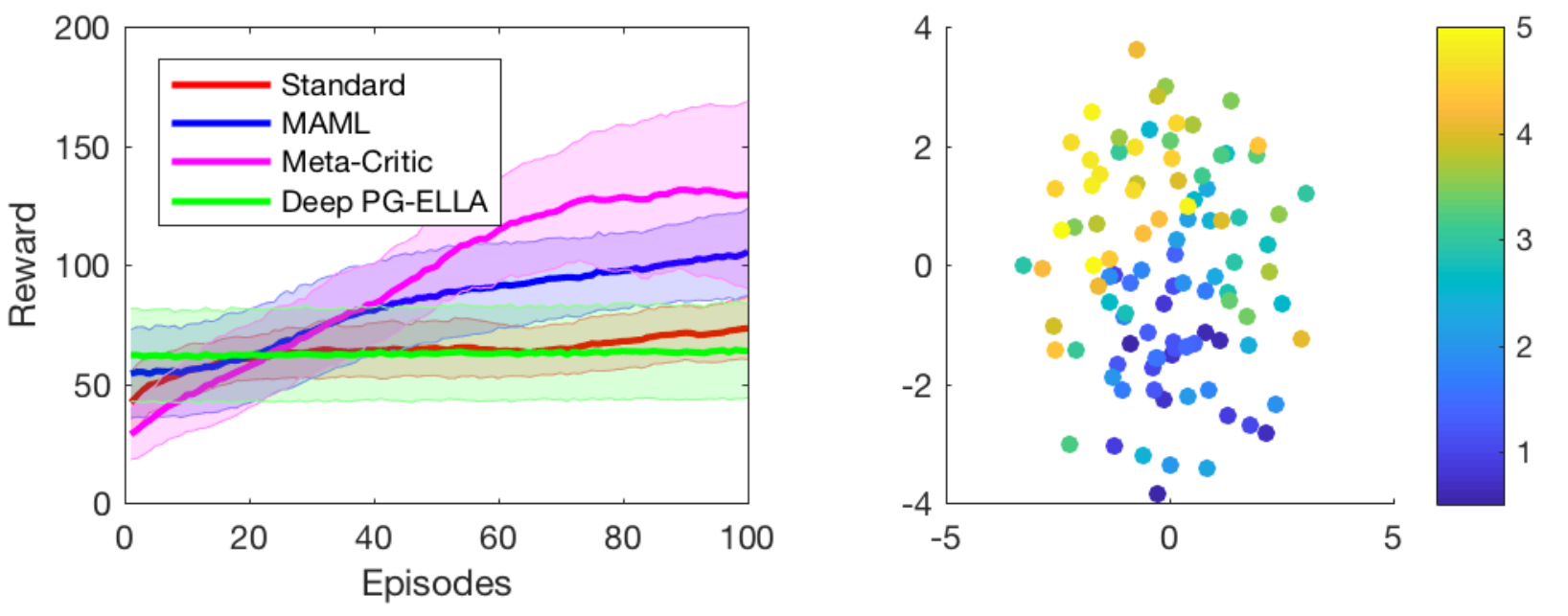}~~
\includegraphics[width=0.29\linewidth]{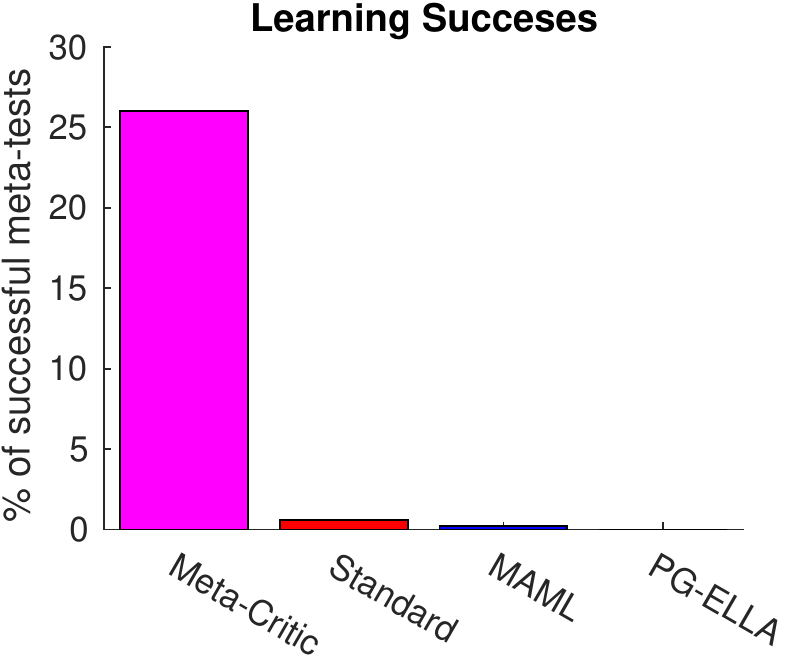}
\caption{{\small Cartpole results. Left: Average reward per episode when learning a new task. Middle: t-SNE plot of cartpole task embeddings. Each point is a task and colour indicates pole length. Right: Percentage of successful meta-tests (learned to consistently balancing a pole of a new length).}}\label{fig:mvn_cartpole}
\end{figure}

The results in Fig.~\ref{fig:mvn_cartpole}(left) show that the meta-critic has a much higher learning rate (improvement per-training episode), and quickly learns to outperform the others. It's starting point is slightly lower, since it starts from a randomly initialised actor policy unlike MAML and PG-ELLA. Quantifying a successful meta-test as succeeding on 10 corresponding final tests (a successful episode is in turn defined in the standard way of balancing the pole for $>195$ iterations \cite{gym}), Fig.~\ref{fig:mvn_cartpole}(right) shows that Meta-Critic's success rate of $26\%$ is much higher than alternatives. Finally, Fig.~\ref{fig:mvn_cartpole}(middle) shows a t-SNE \cite{ictdbid:2777} plot of the TAEN embeddings $z$ coloured by the pole length of the task. The learned embedding clearly reflects a progression of pole length. This is noteworthy as the agent is not directly exposed to the pole length. It has learned this task manifold based only on learning traces.

\section{Conclusion}

We have presented a very flexible meta-learning method for few-shot learning that applies to both reinforcement and supervised learning settings, and can seamlessly exploit unlabelled data. Promising results are obtained on a variety of problems. In future work we would like to evaluate our method to more challenging problems in supervised learning and RL for control in terms of difficulty and input/output dimensionality; and extend it to the continual learning setting.

\bibliographystyle{abbrvnat}

\end{document}